\documentclass[10pt,twocolumn,letterpaper]{article}
\usepackage{cvpr}  
\usepackage[dvipsnames]{xcolor}

\usepackage{color}
\usepackage[dvipsnames]{xcolor}
\usepackage{multirow}
\usepackage{amsmath}
\input{def.set}
\usepackage{colortbl}
\definecolor{Gray}{gray}{0.9}
\definecolor{cvprblue}{rgb}{0.21,0.49,0.74}
\usepackage[pagebackref,breaklinks,colorlinks,citecolor=cvprblue]{hyperref}
\usepackage[accsupp]{axessibility}

\def\confName{CVPR}
\def\confYear{2024}
\title{Boosting Diffusion Models with Moving Average Sampling in Frequency Domain\thanks{This work was performed at HiDream.ai.}}

\author{Yurui Qian$^\dag$, Qi Cai$^\ddag$, Yingwei Pan$^\ddag$, Yehao Li$^\ddag$, Ting Yao$^\ddag$, Qibin Sun$^\dag$, and Tao Mei$^\ddag$ \\
	$^\dag$University of Science and Technology of China \quad $^\ddag$HiDream.ai Inc. \\
	{\tt\small qyr123@mail.ustc.edu.cn}, {\tt\small\{cqcaiqi, pandy, liyehao, tiyao\}@hidream.ai},\\ {\tt\small qibinsun@ustc.edu.cn}, {\tt\small tmei@hidream.ai}
}

\begin{document}
	\maketitle
	
	\begin{abstract}
		Diffusion models have recently brought a powerful revolution in image generation. Despite showing impressive generative capabilities, most of these models rely on the current sample to denoise the next one, possibly resulting in denoising instability. In this paper, we reinterpret the iterative denoising process as model optimization and leverage a moving average mechanism to ensemble all the prior samples. Instead of simply applying moving average to the denoised samples at different timesteps, we first map the denoised samples to data space and then perform moving average to avoid distribution shift across timesteps. In view that diffusion models evolve the recovery from low-frequency components to high-frequency details, we further decompose the samples into different frequency components and execute moving average separately on each component. We name the complete approach ``Moving Average Sampling in Frequency domain (MASF)''. MASF could be seamlessly integrated into mainstream pre-trained diffusion models and sampling schedules. Extensive experiments on both unconditional and conditional diffusion models demonstrate that our MASF leads to superior performances compared to the baselines, with almost negligible additional complexity cost.
	\end{abstract}
	
	\section{Introduction}
	
	\begin{figure}[t]
		\centering
		\vspace{-0.13in}
		\includegraphics[width=0.9\linewidth]{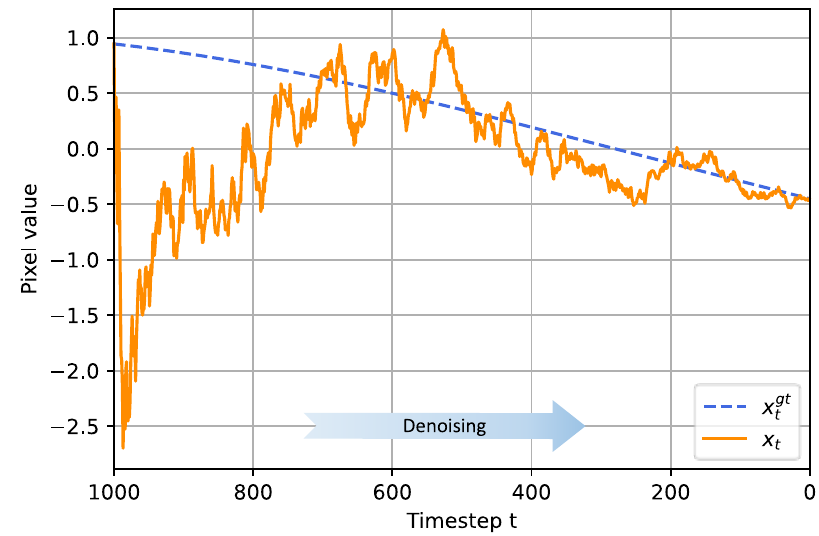}
		\vspace{-0.1in}
		\caption{
			We utilize a diffusion model from ADM \cite{adm} pre-trained on ImageNet-64 to sample from white noise and capture the intermediate output as denoised sample $\bx_t$. Subsequently, we plot the pixel value of $\bx_t$ with respect to generative timesteps. The starting point and ending point of the trajectory of $\bx_t$ are regarded as the ground truth for the noisy image and clean image, respectively. With that we calculate the ground truth of $\bx_t$ at any timestep $t$ using the formulation defined in forward process.}
		\vspace{-0.2in}
		\label{fig:xt_curve}
	\end{figure}
	
	Diffusion model is an increasingly appealing direction to improve the state-of-the-art innovations for generative tasks. In the regime of computer vision, multiple milestones under diffusion models have been established on unconditional and conditional image synthesis \cite{ddpm,ddim,rombach2022high,chen2023controlstyle}, image restoration \cite{xia2023diffir,luo2023refusion,fei2023generative}, inpainting \cite{lugmayr2022repaint,li2017localization,anciukevivcius2023renderdiffusion}, captioning \cite{luo2023semantic}, video generation \cite{ho2022imagen,yang2023diffusion,esser2023structure}, 3D/audio synthesis \cite{chen2023control3d,kong2020diffwave,schneider2023archisound,yang20233dstyle}. In general, the diffusion model consists of a forward process that progressively introduces Gaussian noise into an image, and a denoising network that approximates the reverse of the forward process to produce an image from noise. Compared to generative adversarial networks (GAN) \cite{goodfellow2014generative,karras2019style,pan2017create}, diffusion models are shown capable of better training stability and less sensitivity to hyperparameters, leading to high-quality and coherent samples and alleviating mode collapse. The representative works are DDPM \cite{ddpm} and DDIM \cite{ddim}, which generate samples from white noise by a Markov chain and a ``short'' generative Markov chain corresponding to non-Markovian forward process, respectively. Despite accelerating sampling several orders of magnitude by DDIM, both DDPM and DDIM capitalize only on the current sample to produce the next one, introducing discretization errors \cite{lu2022dpm}. The stochastic nature of these errors brings instability into denoising.
	As illustrated in Figure \ref{fig:xt_curve}, the denoised sample thus tends to oscillate around the ground truth value and the denoising process necessitates lengthy timesteps to converge. To mitigate this issue, some high-order solvers including DPM-Solver \cite{lu2022dpm, dpm-v2} and UniPC \cite{zhao2023unipc}, exploit the resulting samples of the previous $K$ (usually 1 to 3) timesteps to refine the prediction of the current sample. Nevertheless, the prior information in the generative process is still not yet fully leveraged for denoising.

	Another observation in diffusion models is the frequency principle that diffusion models denoise the low-frequency signals first, and gradually add high-frequency details into the sample. Most existing diffusion approaches, however, seldom explore this principle in generative process and treat the sampling consistent as the denoising proceeds, regardless of the connection between frequency evolution and generative timesteps. In contrast, Spectral Diffusion \cite{yang2023diffusionslim} presents dynamic feature extraction at each generative timestep to adjust frequency characteristics. The requirements of a specialized network design and the re-training of the denoising network make it difficult to apply to existing diffusion models.

	In response to the above issues, we propose a training-free approach, namely Moving Average Sampling in Frequency domain (MASF), to enhance the stability of generative process. Technically, MASF reframes the iterative denoising process as model optimization and delves into the moving average mechanism to utilize all the prior samples at each timestep. Note that the diffusion models denoise samples from $\bx_T$ to $\bx_0$, and each sample inherently lies in distinct distribution between white noise and the initial data distribution. This case deviates from the assumption in model optimization that the parameters should come from a constant distribution. As such, it is inappropriate to directly apply moving average to $\bx_t$. Instead, we map sample $\bx_t$ to the data space $\bx_0$ and execute moving average on $\bx_0$ to reduce distribution shift across sampling at different timesteps. Furthermore, MASF decomposes the sample into frequency components and performs separate moving average on each component to dynamically evolve different components along the denoising process. Specifically, we devise a weighting scheme to prioritize low-frequency components denoising in the early timesteps and progressively contribute more weights to high-frequency components in the later timesteps. To ensure compatibility with existing diffusion networks that only accept the complete sample, we reconstruct the sample from all frequency components before feeding it into the denoising network at each timestep. The conversion between sample and frequency components only introduces insignificant extra overheads.

	In summary, we have made the following contributions: 1) The proposed MASF is shown capable of leveraging all~the prior samples in frequency domain to better denoise the current sample in generative process. 2) The exquisitely designed MASF is shown able to be seamlessly integrated into existing diffusion models. 3) MASF has been properly analyzed and verified through extensive experiments on both unconditional and conditional diffusion models to validate its efficacy.

	\section{Related Work}
	\textbf{Sampling Methods in Diffusion Models.}
	Diffusion models have made tremendous progress in generating high-fidelity images \cite{ldm, adm, NEURIPS2022_ec795aea, zhang2023adding, ruiz2023dreambooth}, wherein sampling methods play a crucial role in unleashing their power for image generation with minimal computational cost. \cite{sde} first formulates the diffusion sampling process as the solving of stochastic differential equations (SDEs) \cite{zhang2022gddim, bao2022analytic, jolicoeurmartineau2021gotta, gonzalez2023seeds} and ordinary differential equations (ODEs) \cite{ddim, lu2022dpm, pndm, zhao2023unipc}. Specifically, some works \cite{zhang2022fast, lu2022dpm, dpm-v2, gonzalez2023seeds} build on approximating exponential integrators to reduce the truncation error, while the others \cite{pndm, li2023era} follows traditional numeric methods \cite{rungekutta, wells1982multirate} to solve ODEs. In addition, \cite{adm, ho2021classifierfree} imposes advanced guidance during sampling process to facilitate generation. In this work, we propose an orthogonal design to the aforementioned sampling methods from the perspective of improving denoising stability.

	\noindent \textbf{Sampling Process Stabilization.}
	In an effort to alleviate the denoising instability issue, some prior works explore momentum to stabilize sampling process. For instance, \cite{artifact} examines the divergence artifacts in scenarios with limited sampling steps and incorporates the momentum technique into existing sampling methods. \cite{wu2023fast} associates the diffusion process with stochastic optimization procedure and draws inspiration from momentum SGD to design momentum-based forward process to accelerate training convergence. Similarly, inspired by Adam optimizer, \cite{wang2023boosting} proposes a new sampler that follows the convention in Adam optimizer to define their momentum and update velocity. b   nmIn contrast, our proposed MASF executes moving average in frequency domain to novelly excavate the frequency dynamics for stabilizing sampling process along frequency evolution.

	\noindent \textbf{Frequency Modeling in Diffusion Models.}
	Wavelet decomposition \cite{graps1995introduction, 192463} has been widely adopted in conventional generative methods (e.g., GANs \cite{gal2021swagan, yang2022fregan, Yang2022WaveGAN, zhang2022styleswin}) to exploit additional frequency-aware information in frequency domain. Recently, several advances start to integrate diffusion models with wavelet information. In particular, \cite{li2023colorization} employs score-based models in wavelet spectrum to promote image colorization, while \cite{guth2022score} accelerates score-based generative models by factorizing data distribution into multiscale conditional probabilities of wavelet coefficients. Additionally, \cite{phung2023wavediff} and \cite{yuan2023spatial} design frequency-aware architectures to process data in frequency domain, pursuing faster processing and higher image quality, respectively. Spectral Diffusion \cite{yang2023diffusionslim} also studies the frequency evolution in denoising procedure, and utilizes wavelet gating to trigger spectrum-aware distillation, leading to reduced computation cost. Nevertheless, Spectral Diffusion requires additional re-training of a specialized denoising network, and thus fails to be directly applied to different diffusion models. Instead, our approach seeks a training-free solution that exploits all the prior samples in frequency domain to strengthen the stability of denoising process, which can be seamlessly integrated into any diffusion models.

	\section{Preliminaries}
	Here we first briefly review the typical Denoising Diffusion Probabilistic Models (DDPM) \cite{ddpm} and Denoising Diffusion Implicit Model (DDIM) \cite{ddim} for sampling.
	
	\noindent \textbf{Denoising Diffusion Probabilistic Models.} DDPM consists of a forward process and a denoising process. For the forward process, DDPM transitions from intractable data distribution, denoted as $\bx_0 \sim q_0(\bx_0)$, to Gaussian distribution $q_T(\bx_T) \sim \mathcal{N}(\bx_T; \bzero, \bI)$. This is achieved by progressively adding Gaussian noise to the original image $\bx_0$, and the transition distribution at timestep $t$ is thus defined as:
	\begin{align}
		q(\bx_t| \bx_{t-1} ) & = \mathcal{N} (\bx_t; \sqrt{\alpha_t} \bx_{t-1}, (1 - \alpha_t)\bI),
	\end{align}
	where $\alpha_1, \dots, \alpha_T$ are predefined variance schedules. Following the properties of chained Gaussian processes, DDPM defines $\bar{\alpha}_t = \prod_{i=1}^{t} \alpha_i$ and the value of $\bx_t$ is thus calculated in a single step:
	\begin{equation}
		\label{eq:xt and x0}
		\bx_t = \sqrt{\bar{\alpha}_t} \bx_0 + \sqrt{1 - \bar{\alpha}_t} \bepsilon, \bepsilon \sim \mathcal{N}(\bzero, \bI).
	\end{equation}
	The denoising process employs a neural network to approximate the conditional distribution $p_{\theta}(\bx_{t-1} | \bx_t)$. The optimization objective can be derived from the variational lower bound, which is expressed as:
	\begin{equation}
		\label{eq:loss}
		L = D_{KL} ( q(\bx_{t-1}| \bx_t, \bx_0) || p_{\theta}(\bx_{t-1} | \bx_t)).
	\end{equation}
	The analytical form of $q(\bx_{t-1}|\bx_t, \bx_0)$ is defined by $\mathcal{N}(\bx_{t-1}; \tilde{\bmu}_{t-1}, \sigma_{t-1}^2 \bI)$, while $p_{\theta}(\bx_{t-1} | \bx_t) $ takes the form $ \mathcal{N}(\bx_{t-1}; \bmu_{\theta}(\bx_t, t),\sigma_{t-1}^2 \bI)$. The specific formulations of $\tilde{\bmu}_{t-1}$ and $\bmu_{\theta}(\bx_t, t)$ are measured as follows:
	\begin{align}
		\label{eq:ptheta}
		\tilde{\bmu}_{t-1}      & = \frac{1}{\sqrt{\alpha_t}} (\bx_t - \frac{1 - \alpha_t}{\sqrt{1 - \bar{\alpha}_t}} \bepsilon),                    \\
		\bmu_{\theta}(\bx_t, t) & = \frac{1}{\sqrt{\alpha_t}} (\bx_t - \frac{1 - \alpha_t}{\sqrt{1 - \bar{\alpha}_t}} \bepsilon_{\theta}(\bx_t, t)).
	\end{align}
	In this context, $\bepsilon_{\theta}(\bx_t, t)$, generated by the diffusion model, serves to approximate the added noise $\bepsilon$, based on the noised image $\bx_t$ and the specific timestep $t$. After eliminating constant scaling factors in this loss function, DDPM allows for a simplification of the optimization objective:
	\begin{equation}
		L_{t-1} = \mathbb{E}_{\bx_0, \bepsilon} \left[ \|\bepsilon - \bepsilon_{\theta}(\bx_t, t)\|^2 \right].
	\end{equation}

	\noindent \textbf{Denoising Diffusion Implicit Model.} DDIM upgrades DDPM framework by integrating a non-Markovian process, which effectively decouples $\bx_{t-1}$ from $\bx_t$, enabling the skipping of timesteps to accelerate the sampling process. In this way, DDIM redefines the denoising distribution as:
	\begin{align}\scriptsize
		p_{\theta}(\bx_{t-1} | \bx_t) = \mathcal{N}(\sqrt{\bar{\alpha}_{t-1}} \bx_0^{t} + \sqrt{1 - \bar{\alpha}_t - \eta_t^2} \frac{\bx_t - \sqrt{\bar{\alpha}_t} \bx_0^{t} }{\sqrt{1 - \bar{\alpha}_t}}, \eta_t^2 \bI). \label{eq:ddim ptheta}
	\end{align}
	Here $\bx_0^{t}$ denotes the estimated original sample from the perturbed sample $\bx_t$, which is calculated as:
	\begin{align}\small\label{eq:x0_pred}
		\bx_0^{t} = (\bx_t - \sqrt{1 - \bar{\alpha}_t} \bepsilon_{\theta}(\bx_t,t) )/\sqrt{\bar{\alpha}_t}.
	\end{align}
	The model's behavior is contingent on the value of $\eta_t$. Specifically, when $\eta_t$ is set as $\sqrt{(1- \bar{\alpha}_{t-1}) / (1 - \bar{\alpha}_t)} \sqrt{1 - \alpha_t}$, it aligns with the sampling process of DDPM. Conversely, setting $\eta_t$ to 0 completely eliminates stochasticity during sampling, thereby turning into the sampling process of DDIM.

	\begin{figure}[t]
		\centering
		\includegraphics[width=\linewidth]{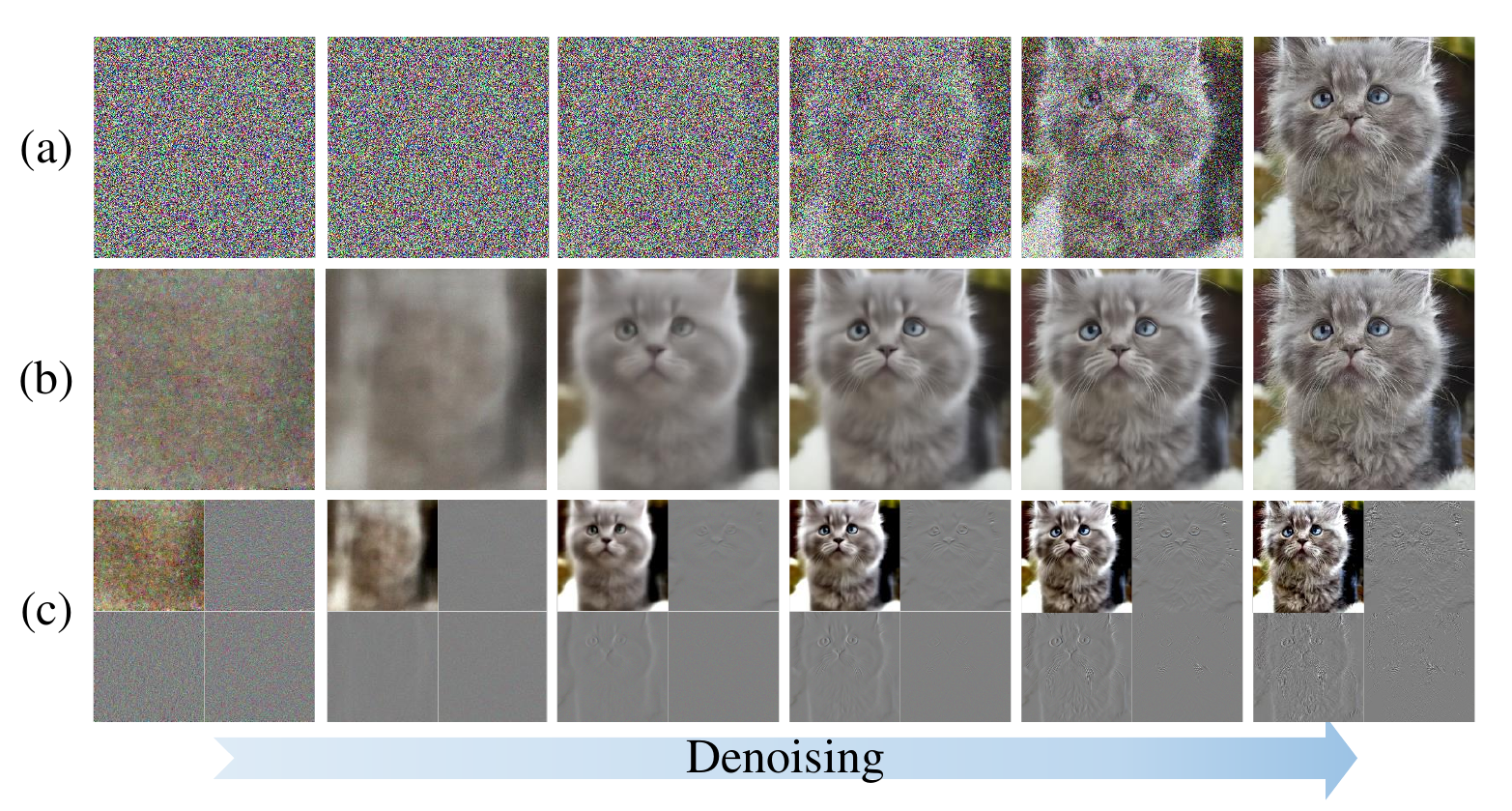}
		\vspace{-0.25in}
		\caption{The evolution of (a) denoised sample $\bx_t$, (b) estimated sample in data space $\bx_0^{t}$ and (c) four subbands in frequency domain of $\bx_0^{t}$ along denoising process. Here each group of subbands is achieved via wavelet decomposition, yielding four different frequency components: $ll$ ($\nwarrow$), $lh$ ($\nearrow$), $hl$ ($\swarrow$), and $hh$ ($\searrow$).}
		\vspace{-0.2in}
		\label{fig:freq-img}
	\end{figure}

	\begin{figure*}[t]
		\centering
		\includegraphics[width=0.95\linewidth]{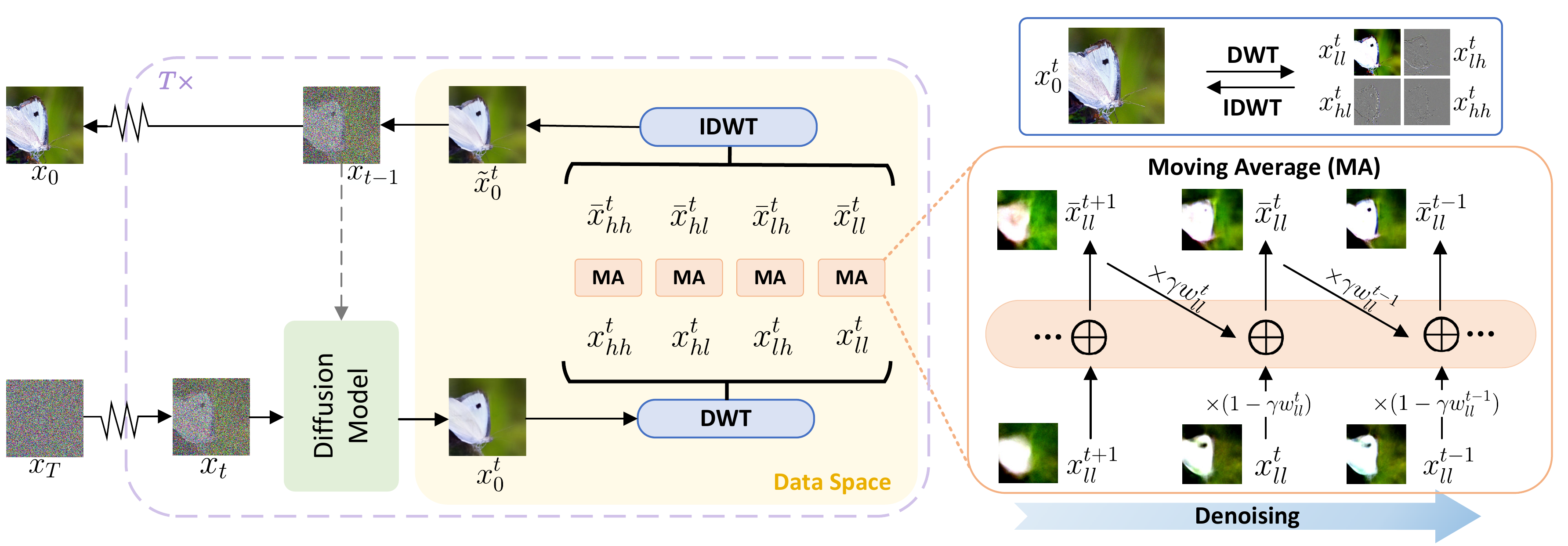}
		\vspace{-0.2in}
		\caption{The overall framework of our Moving Average Sampling in Frequency domain (MASF) for denoising stabilization. At each denoising timestep $t$, MASF first maps the denoised sample $\bx_t$ into data space, leading to the estimated sample $\bx_0^{t}$. We then perform frequency decomposition of $\bx_0^{t}$ via Discrete Wavelet Transformation (DWT) and achieve four subbands ($\bx^t_{\{ll, lh, hl, hh\}}$). After that, MASF updates each frequency component (e.g., the low-frequency component $\bx^t_{ll}$) through moving average over prior samples, pursuing harmonized stabilization along with frequency evolution. The refined subbands $\bar{\bx}^t_{\{ll, lh, hl, hh\}}$ are finally converted back to image domain via Inverse DWT (IDWT) to trigger the subsequent denoising process.}
		\vspace{-0.2in}
		\label{fig:framework}
	\end{figure*}
	
	\section{Our Approach}
	Now we proceed to present our central proposal, Moving Average Sampling in Frequency domain (MASF), aiming to enhance the stability of the denoising process. This section starts by introducing the moving average sampling within the data space. After that, we novelly capitalize on Discrete Wavelet Transformation (DWT) to extend such moving average strategy into the frequency domain. Furthermore, a new dynamic weighting scheme is designed to dynamically perform moving average over different frequency components, pursuing harmonized stabilization along with frequency evolution at denoising process. Figure~\ref{fig:framework} depicts an overview of our MASF framework.
	
	\subsection{Moving Average in Data Space} \label{sec:ama}
	Recall that during typical denoising process, the estimated denoised sample $\bx_t$ always oscillates around its ground truth value due to the stochastic nature of discretization errors \cite{lu2022dpm}. The local errors committed at each timestep accumulate into the global error, which can potentially disrupt the sampling process and result in denoising instability. 
	Intuitively, as shown in Figure \ref{fig:freq-img} (a), the denoising process resembles model optimization, where the denoised sample is iteratively refined via the learnt diffusion model, akin to parameter (denoised sample) optimization (denoising) in model training. This observation motivates us to explore the commonly adopted moving average technique in conventional model optimization, to stabilize the optimization trajectory of the denoised sample $\bx_t$ at inference. Nevertheless, considering that denoised samples $\bx_t$ at different timesteps are perturbed at various noise scales, simply applying moving average over primary $\bx_t$ derived from distinct distributions might inject harmful distortions into optimization trajectory. 
	As an alternative, we propose to map the denoised sample $\bx_t$ back to initial data space, leading to estimated sample $\bx_0^{t}$ consistently predicting clean sample (see Figure \ref{fig:freq-img} (b)).
	After that, we perform moving average over prior estimated samples $\bx_0^t$ to stabilize the denoising process. It is worthy to note that this design of moving average in data space is readily pluggable to any sampling solver. We next discuss how to integrate moving average into DDIM framework \cite{ddim} and other solvers.
	
	\textbf{Moving Average in DDIM Solver.} Since DDIM explicitly computes estimated sample $\bx_0^{t}$ during sampling, our moving average design can be directly applied over the measured $\bx_0^{t}$. Formally, in order to stabilize the optimization trajectory of $\bx_0^t$, we maintain a global moving average $\bar{\bx}_0^t$ that aggregates all prior estimated samples. At each timestep $t$, we update $\bar{\bx}_0^t$ by additionally augmenting the observed $\bx_0^{t}$ ($0 \le t <T$):
	\begin{equation}\small
		\label{eq:ema}
		\bar{\bx}^{t}_0 = ( 1- \gamma ) \bx_0^{t} + \gamma \bar{\bx}_0^{t+1},
	\end{equation}
	where $\gamma$ is a balancing hyperparameter that controls the degree of dependence on the previous moving average versus the current sample. A larger $\gamma$ reflects more reliance on previous $\bar{\bx}_0^{t+1}$ but less focus on $\bx_0^t$, leading to a smoother trajectory. Based on $\bar{\bx}^{t}_0$, we can estimate more stable and accurate denoised sample $\bx_{t-1}$ by simply replacing $\bx_0^t$ with the global moving average $\bar{\bx}^{t}_0$ in Eq. (\ref{eq:ddim ptheta}) of DDIM.
	
	In addition, we observe that the different spatial locations in an image evolve in different rates during denoising process. For example, as shown in Figure \ref{fig:freq-img} (b), the cat's face in $\bx_0^t$ evolves more sharply than left bottom corner of background. Motivated by this, we introduce an adaptive weight $\bw_t$ to modify $\gamma$ for different spatial locations. Here $\bw_t$ is measured as the discrepancy between $\bx_0^{t}$ and $\bar{\bx}_0^{t+1}$. In this way, when the spatial location evolves sharply (i.e., large discrepancy between $\bx_0^{t}$ and $\bar{\bx}_0^{t+1}$), we will amplify the reliance on the global moving average $\bar{\bx}_0^{t+1}$ to pursue more stable denoising process. Eventually, the update function of $\bar{\bx}_0^t$ is operated as:
	\begin{equation}\label{eq:ama}\small
		\bar{\bx}^{t}_0 =( \bone - \gamma \bw_t)\circ \bx_0^{t} +  \gamma  \bw_t \circ \bar{\bx}_0^{t+1},
	\end{equation}
	where $\circ$ denotes element-wise multiplication.

	\textbf{Moving Average in Other Solvers.} For solvers \cite{ddim, dpm-v2} that explicitly define $\bx_0$ in their formulations, we can directly apply Eq. (\ref{eq:ama}) to integrate the moving average technique. 
	For other solvers \cite{ddpm, pndm} which leverage ${\bepsilon}_{\theta}(\bx_t,t)$ instead of $\bx_0$ in updating function, we first calculate $\bx_0$ using Eq. (\ref{eq:x0_pred}) and then apply moving average as in Eq. (\ref{eq:ama}). 
	Subsequently, $\bx_0^t$ is replaced with $\bar{\bx}^{t}_0$ to achieve a refined $\bar{\bepsilon}_{\theta}(\bx_t,t) = (\bx_t - \sqrt{\bar{\alpha}_t} \bar{\bx}_0 ^t)/\sqrt{1 - \bar{\alpha}_t}$, which can be seamlessly integrated into those solvers.
	
	\begin{figure*}[t]
		\vspace{-0.2in}
		\centering
		\begin{subfigure}{0.47\linewidth}
			\includegraphics[width=\linewidth]{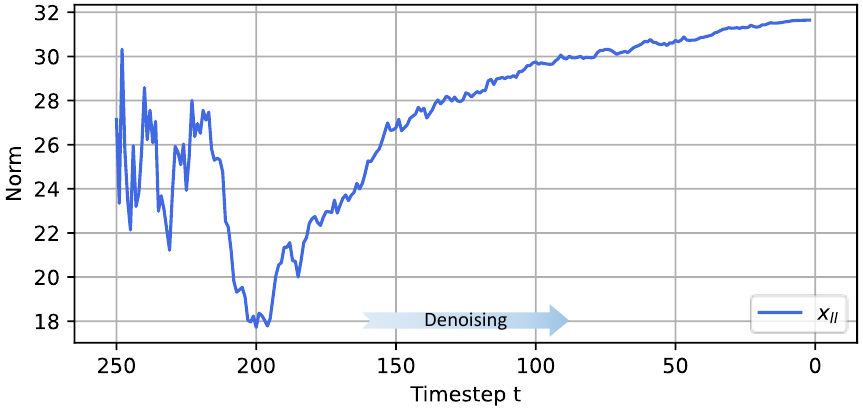}
			\caption{Low-frequency Component.}
		\end{subfigure}
		\begin{subfigure}{0.47\linewidth}
			\includegraphics[width=\linewidth]{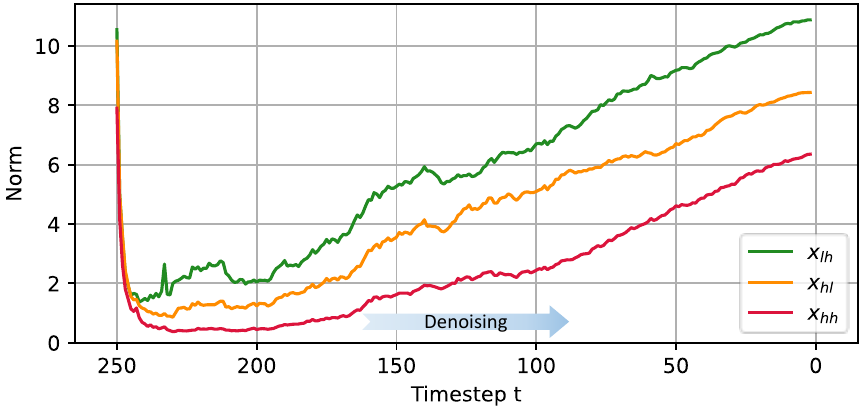}
			\caption{High-frequency Components.}
		\end{subfigure}
		\vspace{-0.1in}
		\caption{Evolution ($l_2$ norm) of different frequency subbands during denoising process. (a) The low-frequency subband oscillates sharply at the beginning and stabilizes after a certain timestep. (b) In contrast, the $l_2$ norm of high-frequency subbands drops rapidly into a small value and then increases steadily along with the denoising process. (We use the model from ADM \cite{adm} pre-trained on ImageNet-64)}
		\vspace{-0.5cm}
		\label{fig:freq-trend}
	\end{figure*}
	
	\subsection{Moving Average in Frequency Domain} \label{sec:freq}
	A well-known evolution law of diffusion models at denoising process is first focusing on the recovery of low-frequency component in the earlier timesteps and gradually turning to recover high-frequency details in the later timesteps. Taking the decomposed frequency components in Figure \ref{fig:freq-img} (c) as an example, the low-frequency component ($ll$) only evolves sharply in the earlier timesteps, while the high-frequency components ($lh$, $hl$, $hh$) start to evolve significantly in the later timesteps. As such, we further extend moving average sampling into frequency domain by executing moving average separately on different frequency components, thereby encouraging more harmonized stabilization along with frequency evolution.

	Technically, MASF employs Discrete Wavelet Transformation (DWT) \cite{graps1995introduction} to decompose the estimated sample $\bx_0^{t} \in \mathbb{R}^{H \times W}$ into four wavelet subbands: $\bx^t_{ll}$, $\bx^t_{lh}$, $\bx^t_{hl}$, $\bx^t_{hh}$. The dimension of each subband is \(\mathbb{R}^{H/2 \times W/2}\). Note that here we implement DWT as the classical Haar wavelet \cite{stankovic2003haar} for simplicity. Among the four wavelet subbands, $\bx^t_{ll}$ refers to the low-frequency component that reflects the basic object structure (resembling a downsampled image), while $\bx^t_{\{lh, hl, hh\}}$ represent high-frequency components that capture texture details. After that, we separately apply moving average over each kind of frequency subband as in Eq. (\ref{eq:ama}):
	\vspace{-0.3cm}
	\begin{equation} \label{eq:freq-ama}
		\small
		\bar{\bx}_f^t = ( \bone - \gamma \bw_f^t) \circ \bx_f^{t} + \gamma \bw_f^t \circ \bar{\bx}_f^{t+1},
	\end{equation}
	where $f \in \{{ll, lh, hl, hh}\}$. As such, each frequency component is independently augmented with the corresponding moving average (denoted as $\bar{\bx}_f^{t}$). This ensures that the trajectories of these frequency components remain non-interfering. Considering that the denoising network is fed with samples in image space, we convert the moving average in frequency domain ($\bar{\bx}_f^t$) back to image domain via Inverse DWT (IDWT) at each timestep.

	\subsection{Frequency Weighting Scheme} \label{sec:freq weight}
	The aforementioned extension of moving average from data space to frequency domain elegantly triggers the interaction between typical denoising process and frequency evolution. Nevertheless, such extension leaves the inherent different priorities for each frequency component along denoising process under-exploited, resulting in the sub-optimal denoising stabilization. In particular, we conduct a comprehensive analysis of frequency evolution in Figure \ref{fig:freq-trend} by visualizing the $l_2$ norm of each frequency component at denoising process. As shown in this figure, the low-frequency subband $\bx^{t}_{ll}$ exhibits sharp oscillation in earlier timesteps and gradually stabilizes after a certain timestep (approximately at 100 steps). Instead, the evolution of high-frequency components $\bx^{t}_f$ ($f \in \{{ll, lh, hl, hh}\}$) reflects a different trend. They drop dramatically into a relatively small value, but subsequently show a steady increase as the denoising process progresses. Such observation reveals that the denoising process often prioritizes reconstructing low-frequency component in the earlier stage, and then focuses on the recovery of high-frequency details later.

	Based on these observations, we further upgrade the moving average in frequency domain with a new dynamic weighting scheme to better align with the evolution dynamics of different frequency components. This dynamic weighting scheme prioritizes low-frequency components in the early timesteps and gradually amplifies the weights of high-frequency components when converting $\bx^{t}_f$ ($f \in \{{ll, lh, hl, hh}\}$) back to image domain. The detailed operation of dynamic weighting scheme is defined as follows:
	\vspace{-0.3cm}
	\begin{equation}
		\small
		\hat{\bx}^t = \text{IDWT}( \beta_f(t) \bx_f^t | f=ll, lh, hl, hh),
	\end{equation}
	where $\beta_{ll} (t)$ decreases linearly as denoising progresses (as timestep evolves from $t+1$ to $t$) and $\beta_{ \{lh, hl, hh \} } (t)$ increase linearly. By integrating previous moving average operation (Eq. (\ref{eq:freq-ama})) with this dynamic weighting scheme, we can achieve the refined version of $\bx_0^t$:
	\vspace{-0.1cm}
	\begin{equation}
		\small
		\tilde{\bx}_0^t = \text{IDWT}\big(\beta_f(t)  \big(( \bone - \gamma \bw_f^t) \circ \bx_f^{t} + \gamma \bw_f^t \circ \bar{\bx}_f^{t+1} \big) \big).
		\vspace{-0.1cm}
	\end{equation}
	Finally, the refined estimated sample $\tilde{\bx}_0^t$ is fed into denoising network to enable a stabilized denoising process.
	
	\begin{table}[t]
		\centering
		\setlength{\tabcolsep}{6pt}
		\caption{FID performances of 50K samples for class-conditional generation on ImageNet with different resolutions and NFEs.}
		\begin{tabular}{lccccc}
			\toprule
			\multicolumn{1}{c}{\multirow{2}{*}{Method}} & \multicolumn{1}{c}{\multirow{2}{*}{Resolution}} & \multicolumn{4}{c}{NFE}                                                    \\ \cline{3-6}
			\multicolumn{1}{c}{}                        & \multicolumn{1}{c}{}                            & 10                      & 15             & 20             & 25             \\
			\midrule
			DDIM                                        & 64                                              & 52.19                   & 24.56          & 15.18          & 11.04          \\
			\rowcolor{Gray}	+MASF                       & 64                                              & \textbf{22.63}          & \textbf{11.23} & \textbf{7.64}  & \textbf{6.32}  \\
			\midrule
			DDIM                                        & 128                                             & 20.36                   & 14.87          & 12.63          & 11.48          \\
			\rowcolor{Gray}	+MASF                       & 128                                             & \textbf{17.19}          & \textbf{12.22} & \textbf{10.16} & \textbf{9.37}  \\
			\midrule
			DDIM                                        & 256                                             & 25.68                   & 19.49          & 17.23          & 16.56          \\
			\rowcolor{Gray}	+MASF                       & 256                                             & \textbf{22.64}          & \textbf{17.67} & \textbf{16.08} & \textbf{15.51} \\
			\bottomrule
		\end{tabular}
		\vspace{-0.2cm}
		\label{tab:main}
	\end{table}

	\section{Experiments}
	We empirically verify the merit of MASF for image generation using diffusion models. The first experiment validates MASF on both conditional and unconditional models across different datasets. The second experiment integrates MASF into recent advances of sampling techniques to examine its impact when combining with state-of-the-art models. The third experiment analyzes how each design in MASF influences the overall performance.

	\begin{table}[t]
		\centering
		\setlength{\tabcolsep}{6.5pt}
		\caption{FID performances of 30K samples for text-conditional generation on MS-COCO with different solvers and NFEs.}
		\begin{tabular}{lcccc}
			\toprule
			\multicolumn{1}{c}{\multirow{2}{*}{Method}} & \multicolumn{4}{c}{NFE}                                                    \\ \cline{2-5}
			\multicolumn{1}{c}{}                        & 10                      & 15             & 20             & 25             \\
			\bottomrule
			DDIM                                        & 35.10                   & 31.05          & 29.18          & 28.51          \\
			\rowcolor{Gray}	+MASF                       & \textbf{25.67}          & \textbf{23.04} & \textbf{22.00} & \textbf{21.80} \\
			\midrule
			DPM-Solver++ \footnotesize (3Fast)    \hspace{-0.4cm}                            & 6.35                    & 6.03           & 6.03           & 5.76           \\
			\rowcolor{Gray} +MASF                       & \textbf{6.20}           & \textbf{6.01}  & \textbf{5.97}  & \textbf{5.69}  \\
			\bottomrule
		\end{tabular}
		\label{tab:coco}
		\vspace{-0.5cm}
	\end{table}

	\subsection{Results on Conditional/Unconditional Models}\label{sec:cond sample}
	\vspace{-0.1cm}
	\textbf{Conditional Models.} Conditional generation leverages control signals into image generation process. The typical conditional models are briefly grouped into two categories: class-conditional and text-conditional. For class-conditional sampling, we utilize pixel-space pre-trained models by the ADM framework \cite{adm} to sample 50K images on different resolutions in the range of 64, 128, and 256, and conduct the experiments on ImageNet \cite{deng2009imagenet} which contains 1,000 distinct classes. We also execute the evaluations with respect to the number of function evaluations (NFE). Table \ref{tab:main} summarizes the Fréchet inception distance (FID) \cite{heusel2017gans} performances of applying MASF to DDIM \cite{ddim} with different NFEs on ImageNet. Overall, using MASF consistently exhibits better FID scores across four NFEs on ImageNet with three image resolutions. The performance gain is larger at small NFE where the instability issue is more severe, demonstrating the advantage of MASF to stabilize the sampling process through moving average in the frequency domain. Notably, MASF brings only 0.97\% extra computational cost to the entire sampling process, making the overhead of the deployment of MASF to diffusion models negligible.

\begin{figure}[t]
		\centering
		\includegraphics[width=\linewidth]{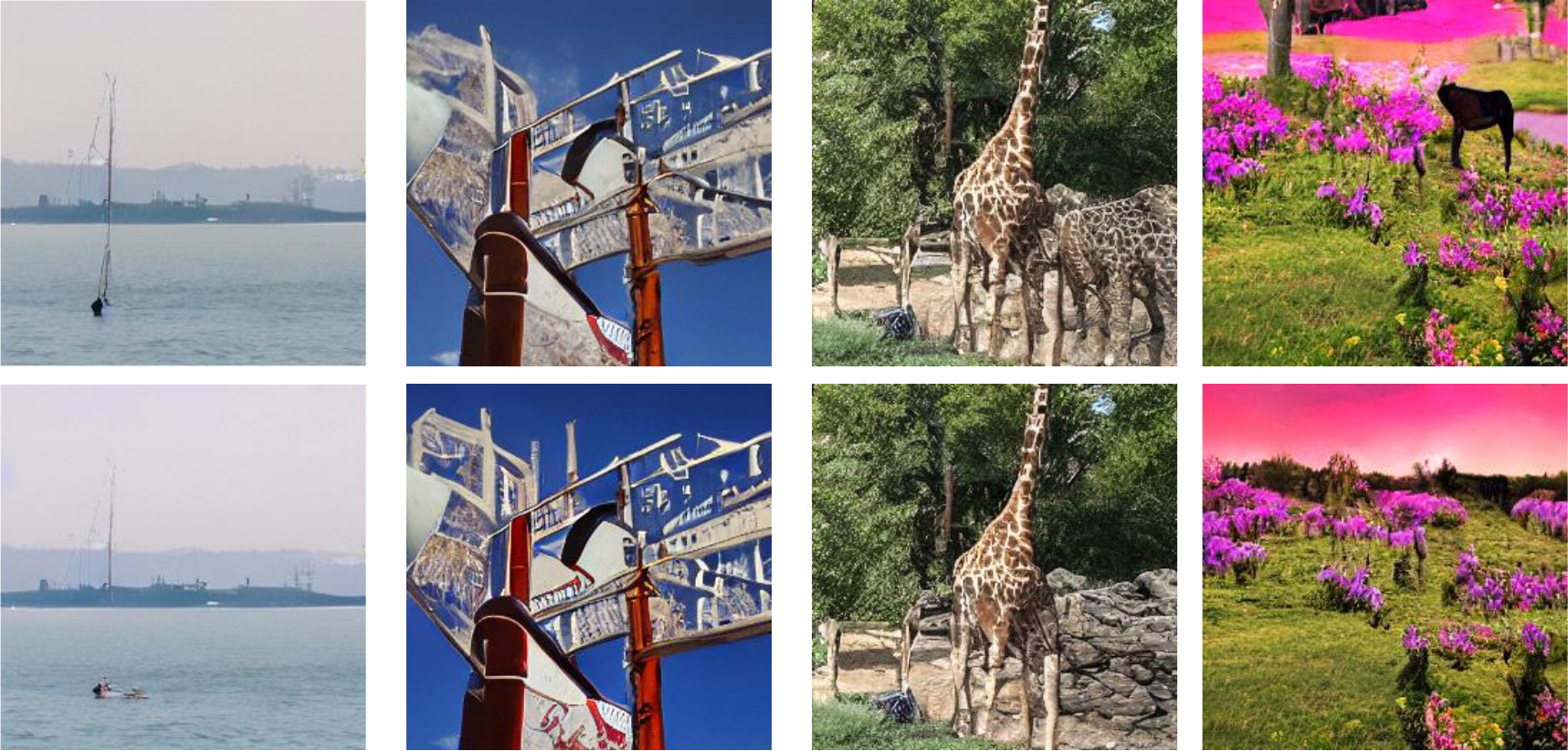}
		\caption{The generated images on MS-COCO using DPM-Solver++ (top) and DPM-Solver++ plus MASF (bottom). }
		\label{fig:coco}
		\vspace{-0.7cm}
	\end{figure}

	For text-conditional generation, we exploit the pre-learnt diffusion model of U-ViT \cite{uvit} on MS-COCO \cite{lin2014microsoft} dataset to produce image samples with the resolution of 256$\times$256. Following the U-ViT protocol, we randomly select 30K prompts from MS-COCO validation set for FID evaluation. Table \ref{tab:coco} lists FID comparisons for text-conditional generation with different solvers and NFEs. As indicated by the results, applying MASF to DDIM reduces the FID score from 35.10 to 25.67 at the NFE of 10, making an absolute improvement of 9.43. Notably, the FID scores on MS-COCO are generally higher than that on ImageNet. We speculate that this may be the result of more complex structures in MS-COCO images. Furthermore, we employ a stronger solver DPM-Solver++ \cite{dpm-v2} instead of DDIM, significantly lowering the FID scores. Impressively, MASF still exhibits its superiority over DPM-Solver++ across all NFEs and MASF leads FID score by 0.07 when sampling 25 steps. To qualitatively validate our MASF, we showcase four image examples generated by DPM-Solver++ and DPM-Solver++ plus MASF in Figure \ref{fig:coco}. The images clearly show that MASF by involving the utilization of moving average in the frequency domain generates higher quality images with less distortions.
	
	\textbf{Unconditional Models.} Different from conditional generation which leverages some specific control signals, unconditional sampling produces images for a particular class completely from pure Gaussian noise. We assess impact of MASF by sampling 50K samples on two widely-adopted datasets: LSUN \cite{lsun} and FFHQ \cite{karras2019style}. For the LSUN dataset, we exploit three pre-trained pixel-space models by the ADM framework \cite{adm} for generating 256$\times$256 images of Horse, Bedroom and Cat, respectively. We run these models with 25 NFE. Table \ref{tab:lsun} details per-class FID performances of unconditional generation on LSUN dataset. Again, DDIM plus MASF constantly improves the FID scores across all three categories. For the extreme case of Bedroom class, DDIM already achieves a very competitive FID score of 4.11, but our MASF still manages to decrease the score to 3.76. For the FFHQ face dataset, we use a latent-space DPM \cite{ldm} to sample images for FID measure. As shown in Table \ref{tab:ffhq}, MASF contributes a FID decrease of 1.82, 0.89, 0.81 and 0.5 with NFE of 10, 15, 20 and 25, respectively, demonstrating the benefit of MASF.
	
	\begin{table}[t]
		\centering
		\setlength{\tabcolsep}{14pt}
		\caption{Per-class FID performances of 50K samples for unconditional generation on LSUN dataset.}
		\vspace{-0.1cm}
		\begin{tabular}{lccc}
			\toprule
			Method                & Horse          & Bedroom       & Cat            \\
			\midrule
			DDIM                  & 18.41          & 4.11          & 12.42          \\
			\rowcolor{Gray} +MASF & \textbf{16.07} & \textbf{3.76} & \textbf{11.65} \\
			\bottomrule
		\end{tabular}
		\label{tab:lsun}
		\vspace{-0.2cm}
	\end{table}

	\begin{table}
		\centering
		\setlength{\tabcolsep}{13pt}
		\caption{FID comparisons of 50K samples for unconditional generation on FFHQ dataset with different NFEs.}
		\vspace{-0.1cm}
		\begin{tabular}{lcccc}
			\toprule
			\multicolumn{1}{c}{\multirow{2}{*}{Method}} & \multicolumn{4}{c}{NFE}                                                 \\ \cline{2-5}
			\multicolumn{1}{c}{}                        & 10                      & 15            & 20            & 25            \\
			\bottomrule
			DDIM                                        & 11.92                   & 6.92          & 5.85          & 5.67          \\
			\rowcolor{Gray} 	+MASF                      & \textbf{10.10}          & \textbf{6.03} & \textbf{5.04} & \textbf{5.17} \\
			\bottomrule
		\end{tabular}
		\vspace{-0.3cm}
		\label{tab:ffhq}
	\end{table}
	\vspace{-0.1cm}
	\subsection{Integration with Other Sampling Techniques} 
	\vspace{-0.1cm}
	\label{sec:classifier-g}
	To further verify the generalizability and effectiveness of MASF on recent advances of diffusion models, we integrate MASF into different sampling techniques including Classifier Guidance \cite{adm} and high-order solvers \cite{dpm-v2,zhao2023unipc}. All the performances here are computed on 50K sampled images with the resolution of 128$\times$128.

	\textbf{Classifier Guidance.}
	As introduced in ADM \cite{adm}, Classifier Guidance enhances generation via modeling the conditional probability of images given a class by a pre-learnt classifier. The diffusion model can take the gradients of the classifier as the condition and a scale is used to adjust the magnitude of the gradients. Table \ref{tab:classifier} lists the FID comparisons with respect to different levels of scales and NFEs. MASF always leads to an FID decrease across all the scales and NFEs. As expected, the scale 0 implies that Classifier Guidance is not involved in this case, yielding inferior performances. The gains of FID scores are between 1.32 and 2.32 at the NFE of 10, when the scale is set in the range of 0.5, 1.0, 2.0 and 4.0. In particular, DDIM plus MASF with guidance scale of 1.0 and NFE of 25 attains the best FID score of 4.73. The results basically demonstrate the effectiveness of our MASF on the guided diffusion models.

	\begin{table}[t]
		\centering
		\setlength{\tabcolsep}{8pt}
		\caption{FID comparisons of 50K samples for conditional generation on ImageNet 128$\times$128 with different guidance scales.}
		\vspace{-0.1cm}
		\begin{tabular}{lccccc}
			\toprule
			\multicolumn{1}{c}{\multirow{2}{*}{Method}} & \multicolumn{1}{c}{\multirow{2}{*}{Scale}} & \multicolumn{4}{c}{NFE}                                                    \\ \cline{3-6}
			\multicolumn{1}{c}{}                        & \multicolumn{1}{c}{}                       & 10                      & 15             & 20             & 25             \\
			\midrule
			
			DDIM                                        & 0.0                                        & 20.36                   & 14.87          & 12.63          & 11.48          \\
			\rowcolor{Gray}	+MASF                       & 0.0                                        & \textbf{17.19}          & \textbf{12.22} & \textbf{10.16} & \textbf{9.37}  \\	\midrule
			
			DDIM                                        & 0.5                                        & 13.61                   & 9.60           & 8.01           & 7.16           \\
			\rowcolor{Gray} 	+MASF                      & 0.5                                        & \textbf{11.29}          & \textbf{7.63}  & \textbf{6.17}  & \textbf{ 5.49} \\
			\midrule
			DDIM                                        & 1.0                                        & 11.18                   & 8.07           & 6.86           & 6.14           \\
			\rowcolor{Gray} 	+MASF                      & 1.0                                        & \textbf{9.39}           & \textbf{6.53}  & \textbf{5.32}  & \textbf{4.73}  \\
			\midrule
			DDIM                                        & 2.0                                        & 10.35                   & 8.11           & 7.19           & 6.66           \\
			\rowcolor{Gray} 	+MASF                      & 2.0                                        & \textbf{8.95}           & \textbf{ 6.82} & \textbf{ 5.92} & \textbf{5.42}  \\
			\midrule
			DDIM                                        & 4.0                                        & 11.61                   & 10.01          & 9.26          & 8.94         \\
			\rowcolor{Gray} 	+MASF                      & 4.0                                        & \textbf{10.29}          & \textbf{8.73} & \textbf{8.06} & \textbf{7.72}  \\
			\bottomrule
		\end{tabular}
		\label{tab:classifier}
		\vspace{-0.3cm}
	\end{table}

	\textbf{High-order Solvers.}
	Next, we extend the evaluation of MASF from on the basic solvers of DDPM and DDIM to high-order solvers of DPM Solver++ \cite{dpm-v2}, UniPC \cite{zhao2023unipc}, and F-PNDM \cite{pndm}. The evaluation protocol follows Classifier Guidance \cite{adm} with a fixed scale of 1.0. Table \ref{tab:solvers} shows the performance comparisons across different solvers. In general, high-order solvers are superior to basic ones, particularly at small NFE. Similar to the observations in Classifier Guidance, integrating MASF into these high-order solvers further improves FID scores, boosting generation quality. The results further validate the design of moving average in frequency domain in our MASF.

	\begin{table}
		\centering
		\setlength{\tabcolsep}{9pt}
		\caption{FID comparisons of 50K samples with different solvers on ImageNet 128$\times$128. * F-PNDM needs at least 12 NFE and is not applicable when NFE=10.}
		\vspace{-0.2cm}
		\begin{tabular}{lcccc}
			\toprule
			\multicolumn{1}{c}{\multirow{2}{*}{Method}} & \multicolumn{4}{c}{NFE}                                                  \\ \cline{2-5}
			\multicolumn{1}{c}{}                        & 10                      & 15             & 20            & 25            \\
			\midrule
			DDPM                                        & 25.24                   & 14.74          & 10.38         & 8.37          \\
			\rowcolor{Gray} 	+MASF                      & \textbf{19.03}          & \textbf{10.60} & \textbf{7.33} & \textbf{5.87} \\
			\midrule
			DDIM                                        & 11.18                   & 8.07           & 6.86          & 6.14          \\
			\rowcolor{Gray} 	+MASF                      & \textbf{9.39}           & \textbf{6.53}  & \textbf{5.32} & \textbf{4.73} \\
			\midrule
			DPM-Solver++\footnotesize(2M)  \hspace{-0.5cm}                              & 5.45                    & 4.54           & 4.18          & 3.98          \\
			\rowcolor{Gray} 	+MASF                      & \textbf{5.25}           & \textbf{4.30}  & \textbf{3.96} & \textbf{3.79} \\
			\midrule
			UniPC                                       & 6.61                    & 4.39           & 4.12          & 3.95          \\
			\rowcolor{Gray} 	+MASF                      & \textbf{6.26}           & \textbf{3.94}  & \textbf{3.68} & \textbf{3.50} \\
			\midrule
			F-PNDM                                      & *                       & 5.98           & 4.56          & 3.30          \\
			\rowcolor{Gray} 	+MASF                      & *                       & \textbf{5.60}  & \textbf{4.13} & \textbf{3.26} \\
			\bottomrule
		\end{tabular}
		\label{tab:solvers}
		\vspace{-0.4cm}
	\end{table}
	
	\subsection{Studies of MASF Designs}
	We perform ablation studies to examine each component's role in MASF. Moreover, we evaluate different $\gamma$ values in moving average, various formulations of spatial weighting $\bw_t$ and frequency weighting $\beta_f(t)$. In view that sampling 50K images is computationally expensive, we generate 10K ImageNet samples of resolution 128$\times$128 with pre-trained model by ADM to do more ablations here.
	
	\textbf{Effect of Each Component.}
	We first study how each particular design in MASF influences the overall performance for image generation. We degrade MASF by removing the frequency domain transformation, termed ``+MA''. Table \ref{tab:components} summarizes the FID improvements by considering one more component at each stage. The ``+MA'' variant applies moving average in pixel space to stabilize the denoising process to some extend, thereby outperforming the base model of DDIM. When further connecting generative sampling and frequency evolution, MASF manifests an apparent FID boost. The results basically prove the complementarity between moving average and frequency domain transformation.

	\begin{table}
		\centering
		\setlength{\tabcolsep}{10pt}
		\caption{FID comparisons of 10K samples with different components on ImageNet 128$\times$128.}
		\begin{tabular}{lcccc}
			\toprule
			\multicolumn{1}{c}{\multirow{2}{*}{Method}} & \multicolumn{4}{c}{NFE}                                                    \\ \cline{2-5}
			\multicolumn{1}{c}{}                        & 10                      & 15             & 20             & 25             \\
			\midrule
			DDIM                                        & 23.00                   & 17.97          & 15.73          & 14.70          \\
			+MA                                         & 21.88                   & 17.48          & 15.42          & 14.58          \\
			\rowcolor{Gray}   +MASF      &    \textbf{21.76}          & \textbf{16.11} & \textbf{13.70} & \textbf{12.80} \\
			
			\bottomrule
		\end{tabular}
		\label{tab:components}
	\end{table}
	
\begin{figure}[t]
		\centering
		\vspace{-0.1cm}
		\includegraphics[width=0.85\linewidth]{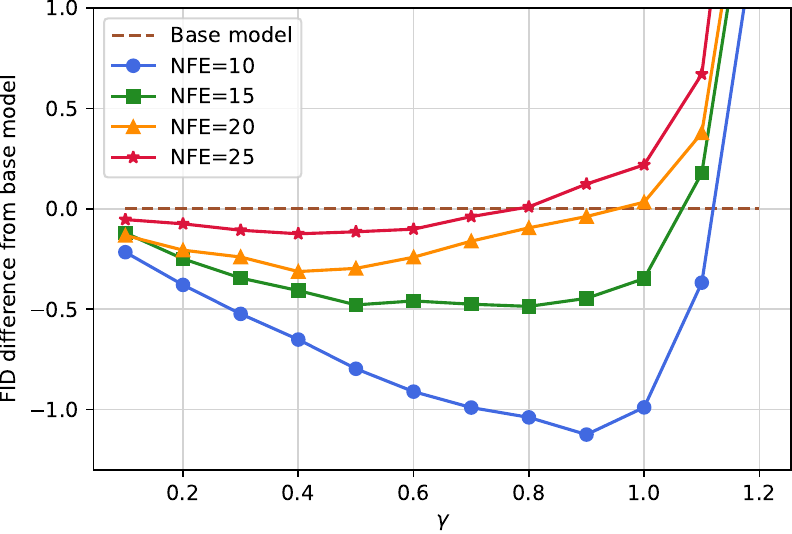}
		\vspace{-0.2cm}
		\caption{FID comparisons of 10K samples with different $\gamma$ on ImageNet 128$\times$128. The y-axis quantifies the FID improvements over the baseline model. Each curve corresponds to a distinct NFE.}
		\label{fig:gammas}
		\vspace{-0.7cm}
	\end{figure}

	\textbf{Effect of $\gamma$ in Moving Average.}
	Next, we test the effect of $\gamma$ in Eq. (\ref{eq:ama}), which balances the contributions of the moving average term and the current predicted sample. In general, a higher value of  $\gamma$ emphasizes more on the moving average term, resulting in a smoother trajectory. In contrast, a lower $\gamma$ value prioritizes the current predicted sample, with $\gamma=0$ downgrading to the base model. Figure \ref{fig:gammas} depicts the FID improvements over the base model through pixel space moving average. An observation is that when the values of $\gamma$ range from 0.1 to 0.8, utilizing moving average consistently yields superior FID scores compared to the base model. Such trend verifies the model's robustness to the variations of $\gamma$ values. Taking a closer look at the curves on different NFEs, a smaller NFE favors a larger $\gamma$, indicating a greater reliance on the moving average term. This observation elegantly corroborates our hypothesis that shorter sampling processes are more vulnerable to instability issues and benefit more from moving average.
	
	\textbf{Effect of $\bw_t$ for Adaptive Weighting.}
	
	In order to analyze the impact of adaptive weight $\bw_t$ in Eq. (\ref{eq:ama}), we further compare different formulations of $\bw_f^t$: Constant weight $\bw_f^t=\bone$, Linear weight $\bw_f^t = |\bx_f^t - \bar{\bx}_f^{t+1}|$ and Quadratic weight $\bw_f^t=|\bx_f^t - \bar{\bx}_f^{t+1}|^2$. The results in Table \ref{tab:bw} indicate that employing Linear weight is superior to Constant weight. The observation supports the spirit behind that when the discrepancy between $\bx_f^t$ and $\bar{\bx}_f^{t+1}$ is large, relying more on moving average makes the denoising process more stable, therefore leading to lower FID. Given the fact that $|\bx_f^t - \bar{\bx}_f^{t+1}|$ is generally smaller than 1, Quadratic weight is smaller than Linear weight, and obtains inferior performances. We speculate that this might be the result of over-reliance on the current predicted sample, making the denoising process less~stable.

	\begin{table}
		\centering
		\setlength{\tabcolsep}{7pt}
		\caption{FID comparisons of 10K samples with different adaptive weight $\bw_t$ on ImageNet 128$\times$128.}
		\begin{tabular}{lcccc}
			\toprule
			\multicolumn{1}{c}{\multirow{2}{*}{Method}}            & \multicolumn{4}{c}{NFE}                                                    \\ \cline{2-5}
			\multicolumn{1}{c}{}                                   & 10                      & 15             & 20             & 25             \\
			\midrule
			$\bw_t=\bone$                                          & 22.05                   & 16.24          & 13.84          & 12.85          \\
			\rowcolor{Gray} 	$\bw_t=|\bx_0^t - \bar{\bx}_0^{t+1}|$ & \textbf{21.76}          & \textbf{16.11} & \textbf{13.70} & \textbf{12.80} \\
			$\bw_t=|\bx_0^t - \bar{\bx}_0^{t+1}|^2$                & 22.06                   & 16.21          & 13.83          & 12.84          \\
			\bottomrule
		\end{tabular}
		\label{tab:bw}
		\vspace{-0.2cm}
	\end{table}
	
	\textbf{Effect of $\beta_f(t)$ for Frequency Weighting.}
	To verify how the frequency weighting scheme $\beta_f(t)$ influences the denoising process, we detail the FID metric with different $\beta_f(t)$ variants. Based on the analysis in Section \ref{sec:freq weight}, frequency component weighting can follow two main trends: linear increase ($\nearrow$) or decrease ($\searrow$) as denoising progresses. The performances are summarized in Table \ref{tab:f-trend}. We have {\small $\beta_{ll}(t_{start})=1.03, \beta_{ll}(t_{end})=1, \beta_{h*}(t_{start})=1, \beta_{h*}(t_{end})=1.13$} for the last row in Table \ref{tab:f-trend}. Either decreasing the weight of low-frequency components (the second row) or increasing the weight of high-frequency components (the fourth row) in denoising process leads to notable FID improvements. Combining them together (the fifth row) yields the most favorable results. This aligns with our analysis of prioritizing low-frequency components in the early timesteps and gradually shifting focus to high-frequency components later in the process.
	\vspace{-0.1cm}
	
	\begin{table}
		\centering
		\caption{FID comparisons of 10K samples with different frequency weight $\beta_f(t)$ on ImageNet 128$\times$128.}
		\vspace{-0.2cm}
		\begin{tabular}{lcccc}
			\toprule
			\multicolumn{1}{c}{\multirow{2}{*}{Method}}       & \multicolumn{4}{c}{NFE}                                                    \\ \cline{2-5}
			\multicolumn{1}{c}{}                              & 10                      & 15             & 20             & 25             \\
			\midrule
			Low $\nearrow$                                    & 25.49                   & 21.08          & 19.27          & 18.34          \\
			Low $\searrow$                                    & {22.80}                 & 17.00          & 14.42          & 13.42          \\
			\midrule
			High $\searrow$                                   & 27.18                   & 20.43          & 20.43          & 19.63          \\
			High $\nearrow$                                   & {21.96}                 & 16.15          & {13.74}        & 12.84          \\
			\midrule
			\rowcolor{Gray} High $\nearrow$ + Low $\searrow$       & 
			\textbf{21.76}          & \textbf{16.11} & \textbf{13.70} & \textbf{12.80} \\
			\bottomrule
		\end{tabular}
		\label{tab:f-trend}
		\vspace{-0.6cm}
	\end{table}
	
	\section{Conclusion}
	\vspace{-0.1cm}
	We have presented the Moving Average Sampling in Frequency domain (MASF), a new technique to enhance the stability of the diffusion process. MASF capitalizes on the moving average mechanism, effectively harnessing all previous samples. Moreover, MASF decomposes the sample into distinct frequency components, allowing for the dynamic evolution of each component during the denoising process. Extensive experiments validate that MASF significantly improves performance across various datasets, models, and sampling techniques. More remarkably, MASF introduces negligible computational overhead and can be readily integrated into existing diffusion models.
	
	{
		\small
		\bibliographystyle{ieeenat_fullname}
		\bibliography{ref.bib}

\begin{thebibliography}{60}
\providecommand{\natexlab}[1]{#1}
\providecommand{\url}[1]{\texttt{#1}}
\expandafter\ifx\csname urlstyle\endcsname\relax
  \providecommand{\doi}[1]{doi: #1}\else
  \providecommand{\doi}{doi: \begingroup \urlstyle{rm}\Url}\fi

\bibitem[Anciukevi{\v{c}}ius et~al.(2023)Anciukevi{\v{c}}ius, Xu, Fisher, Henderson, Bilen, Mitra, and Guerrero]{anciukevivcius2023renderdiffusion}
Titas Anciukevi{\v{c}}ius, Zexiang Xu, Matthew Fisher, Paul Henderson, Hakan Bilen, Niloy~J Mitra, and Paul Guerrero.
\newblock {Renderdiffusion: Image Diffusion for 3D Reconstruction, inpainting and generation}.
\newblock In \emph{CVPR}, 2023.

\bibitem[Bao et~al.(2022)Bao, Li, Zhu, and Zhang]{bao2022analytic}
Fan Bao, Chongxuan Li, Jun Zhu, and Bo Zhang.
\newblock {Analytic-DPM: an analytic estimate of the optimal reverse variance in diffusion probabilistic models}.
\newblock In \emph{ICLR}, 2022.

\bibitem[Bao et~al.(2023)Bao, Nie, Xue, Cao, Li, Su, and Zhu]{uvit}
Fan Bao, Shen Nie, Kaiwen Xue, Yue Cao, Chongxuan Li, Hang Su, and Jun Zhu.
\newblock {All are Worth Words: A ViT Backbone for Diffusion Models}.
\newblock In \emph{CVPR}, 2023.

\bibitem[Butcher(1996)]{rungekutta}
John~Charles Butcher.
\newblock {A history of Runge-Kutta methods}.
\newblock \emph{Applied numerical mathematics}, 20\penalty0 (3):\penalty0 247--260, 1996.

\bibitem[Chen et~al.(2023{\natexlab{a}})Chen, Pan, Yao, and Mei]{chen2023controlstyle}
Jingwen Chen, Yingwei Pan, Ting Yao, and Tao Mei.
\newblock Controlstyle: Text-driven stylized image generation using diffusion priors.
\newblock In \emph{ACM Multimedia}, 2023{\natexlab{a}}.

\bibitem[Chen et~al.(2023{\natexlab{b}})Chen, Pan, Li, Yao, and Mei]{chen2023control3d}
Yang Chen, Yingwei Pan, Yehao Li, Ting Yao, and Tao Mei.
\newblock Control3d: Towards controllable text-to-3d generation.
\newblock In \emph{ACM Multimedia}, 2023{\natexlab{b}}.

\bibitem[Deng et~al.(2009)Deng, Dong, Socher, Li, Li, and Fei-Fei]{deng2009imagenet}
Jia Deng, Wei Dong, Richard Socher, Li-Jia Li, Kai Li, and Li Fei-Fei.
\newblock {ImageNet: A large-scale hierarchical image database}.
\newblock In \emph{CVPR}, 2009.

\bibitem[Dhariwal and Nichol(2021)]{adm}
Prafulla Dhariwal and Alexander Nichol.
\newblock {Diffusion Models Beat GANs on Image Synthesis}.
\newblock In \emph{NeurIPS}, 2021.

\bibitem[Esser et~al.(2023)Esser, Chiu, Atighehchian, Granskog, and Germanidis]{esser2023structure}
Patrick Esser, Johnathan Chiu, Parmida Atighehchian, Jonathan Granskog, and Anastasis Germanidis.
\newblock {Structure and Content-Guided Video Synthesis with Diffusion Models}.
\newblock In \emph{ICCV}, 2023.

\bibitem[Fei et~al.(2023)Fei, Lyu, Pan, Zhang, Yang, Luo, Zhang, and Dai]{fei2023generative}
Ben Fei, Zhaoyang Lyu, Liang Pan, Junzhe Zhang, Weidong Yang, Tianyue Luo, Bo Zhang, and Bo Dai.
\newblock {Generative Diffusion Prior for Unified Image Restoration and Enhancement}.
\newblock In \emph{CVPR}, 2023.

\bibitem[Gal et~al.(2021)Gal, Hochberg, Bermano, and Cohen-Or]{gal2021swagan}
Rinon Gal, Dana~Cohen Hochberg, Amit Bermano, and Daniel Cohen-Or.
\newblock {SWAGAN: A style-based wavelet-driven generative model}.
\newblock \emph{ACM Transactions on Graphics (TOG)}, 40\penalty0 (4):\penalty0 1--11, 2021.

\bibitem[Gonzalez et~al.(2023)Gonzalez, Fernandez, Tran, Gherbi, Hajri, and Masmoudi]{gonzalez2023seeds}
Martin Gonzalez, Nelson Fernandez, Thuy Tran, Elies Gherbi, Hatem Hajri, and Nader Masmoudi.
\newblock {SEEDS: Exponential SDE Solvers for Fast High-Quality Sampling from Diffusion Models}.
\newblock In \emph{NeurIPS}, 2023.

\bibitem[Goodfellow et~al.(2014)Goodfellow, Pouget-Abadie, Mirza, Xu, Warde-Farley, Ozair, Courville, and Bengio]{goodfellow2014generative}
Ian Goodfellow, Jean Pouget-Abadie, Mehdi Mirza, Bing Xu, David Warde-Farley, Sherjil Ozair, Aaron Courville, and Yoshua Bengio.
\newblock {Generative Adversarial Nets}.
\newblock In \emph{NeurIPS}, 2014.

\bibitem[Graps(1995)]{graps1995introduction}
Amara Graps.
\newblock {An Introduction to Wavelets}.
\newblock \emph{IEEE computational science and engineering}, 1995.

\bibitem[Guth et~al.(2022)Guth, Coste, De~Bortoli, and Mallat]{guth2022score}
Florentin Guth, Simon Coste, Valentin De~Bortoli, and Stephane Mallat.
\newblock {Wavelet Score-Based Generative Modeling}.
\newblock In \emph{NeurIPS}, 2022.

\bibitem[Heusel et~al.(2017)Heusel, Ramsauer, Unterthiner, Nessler, and Hochreiter]{heusel2017gans}
Martin Heusel, Hubert Ramsauer, Thomas Unterthiner, Bernhard Nessler, and Sepp Hochreiter.
\newblock {GANs trained by a two time-scale update rule converge to a local nash equilibrium}.
\newblock In \emph{NeurIPS}, 2017.

\bibitem[Ho and Salimans(2021)]{ho2021classifierfree}
Jonathan Ho and Tim Salimans.
\newblock {Classifier-Free Diffusion Guidance}.
\newblock In \emph{NeurIPS 2021 Workshop on Deep Generative Models and Downstream Applications}, 2021.

\bibitem[Ho et~al.(2020)Ho, Jain, and Abbeel]{ddpm}
Jonathan Ho, Ajay Jain, and Pieter Abbeel.
\newblock {Denoising Diffusion Probabilistic Models}.
\newblock In \emph{NeurIPS}, 2020.

\bibitem[Ho et~al.(2022)Ho, Chan, Saharia, Whang, Gao, Gritsenko, Kingma, Poole, Norouzi, Fleet, et~al.]{ho2022imagen}
Jonathan Ho, William Chan, Chitwan Saharia, Jay Whang, Ruiqi Gao, Alexey Gritsenko, Diederik~P Kingma, Ben Poole, Mohammad Norouzi, David~J Fleet, et~al.
\newblock {Imagen Video: High Definition Video Generation with Diffusion Models}.
\newblock \emph{arXiv preprint arXiv:2210.02303}, 2022.

\bibitem[Jolicoeur-Martineau et~al.(2021)Jolicoeur-Martineau, Li, Pich{\'e}-Taillefer, Kachman, and Mitliagkas]{jolicoeurmartineau2021gotta}
Alexia Jolicoeur-Martineau, Ke Li, R{\'e}mi Pich{\'e}-Taillefer, Tal Kachman, and Ioannis Mitliagkas.
\newblock {Gotta Go Fast When Generating Data with Score-Based Models}.
\newblock \emph{arXiv preprint arXiv:2105.14080}, 2021.

\bibitem[Karras et~al.(2019)Karras, Laine, and Aila]{karras2019style}
Tero Karras, Samuli Laine, and Timo Aila.
\newblock {A Style-Based Generator Architecture for Generative Adversarial Networks}.
\newblock In \emph{CVPR}, 2019.

\bibitem[Kong et~al.(2021)Kong, Ping, Huang, Zhao, and Catanzaro]{kong2020diffwave}
Zhifeng Kong, Wei Ping, Jiaji Huang, Kexin Zhao, and Bryan Catanzaro.
\newblock {Diffwave: A Versatile Diffusion Model for Audio Synthesis}.
\newblock In \emph{ICLR}, 2021.

\bibitem[Li et~al.(2017)Li, Luo, and Huang]{li2017localization}
Haodong Li, Weiqi Luo, and Jiwu Huang.
\newblock {Localization of Diffusion-Based Inpainting in Digital Images}.
\newblock \emph{IEEE transactions on information forensics and security}, 12\penalty0 (12):\penalty0 3050--3064, 2017.

\bibitem[Li et~al.(2023{\natexlab{a}})Li, Li, Xu, Wang, and Liu]{li2023colorization}
Jin Li, Wanyun Li, Zichen Xu, Yuhao Wang, and Qiegen Liu.
\newblock {Wavelet Transform-Assisted Adaptive Generative Modeling for Colorization}.
\newblock \emph{IEEE Transactions on Multimedia}, 25:\penalty0 4547--4562, 2023{\natexlab{a}}.

\bibitem[Li et~al.(2023{\natexlab{b}})Li, Liu, Chai, Li, and Tan]{li2023era}
Shengmeng Li, Luping Liu, Zenghao Chai, Runnan Li, and Xu Tan.
\newblock {ERA-Solver: Error-Robust Adams Solver for Fast Sampling of Diffusion Probabilistic Models}.
\newblock \emph{arXiv preprint arXiv:2301.12935}, 2023{\natexlab{b}}.

\bibitem[Lin et~al.(2014)Lin, Maire, Belongie, Hays, Perona, Ramanan, Doll{\'a}r, and Zitnick]{lin2014microsoft}
Tsung-Yi Lin, Michael Maire, Serge Belongie, James Hays, Pietro Perona, Deva Ramanan, Piotr Doll{\'a}r, and C~Lawrence Zitnick.
\newblock {Microsoft COCO: Common objects in context}.
\newblock In \emph{ECCV}, 2014.

\bibitem[Liu et~al.(2022)Liu, Ren, Lin, and Zhao]{pndm}
Luping Liu, Yi Ren, Zhijie Lin, and Zhou Zhao.
\newblock {Pseudo Numerical Methods for Diffusion Models on Manifolds}.
\newblock In \emph{ICLR}, 2022.

\bibitem[Lu et~al.(2022{\natexlab{a}})Lu, Zhou, Bao, Chen, Li, and Zhu]{dpm-v2}
Cheng Lu, Yuhao Zhou, Fan Bao, Jianfei Chen, Chongxuan Li, and Jun Zhu.
\newblock {DPM-Solver++: Fast Solver for Guided Sampling of Diffusion Probabilistic Models}.
\newblock \emph{arXiv preprint arXiv:2211.01095}, 2022{\natexlab{a}}.

\bibitem[Lu et~al.(2022{\natexlab{b}})Lu, Zhou, Bao, Chen, Li, and Zhu]{lu2022dpm}
Cheng Lu, Yuhao Zhou, Fan Bao, Jianfei Chen, Chongxuan Li, and Jun Zhu.
\newblock {DPM-Solver: A Fast ODE Solver for Diffusion Probabilistic Model Sampling in Around 10 Steps}.
\newblock In \emph{NeurIPS}, 2022{\natexlab{b}}.

\bibitem[Lugmayr et~al.(2022)Lugmayr, Danelljan, Romero, Yu, Timofte, and Van~Gool]{lugmayr2022repaint}
Andreas Lugmayr, Martin Danelljan, Andres Romero, Fisher Yu, Radu Timofte, and Luc Van~Gool.
\newblock {Repaint: Inpainting using denoising diffusion probabilistic models}.
\newblock In \emph{CVPR}, 2022.

\bibitem[Luo et~al.(2023{\natexlab{a}})Luo, Li, Pan, Yao, Feng, Chao, and Mei]{luo2023semantic}
Jianjie Luo, Yehao Li, Yingwei Pan, Ting Yao, Jianlin Feng, Hongyang Chao, and Tao Mei.
\newblock Semantic-conditional diffusion networks for image captioning.
\newblock In \emph{CVPR}, 2023{\natexlab{a}}.

\bibitem[Luo et~al.(2023{\natexlab{b}})Luo, Gustafsson, Zhao, Sj{\"o}lund, and Sch{\"o}n]{luo2023refusion}
Ziwei Luo, Fredrik~K Gustafsson, Zheng Zhao, Jens Sj{\"o}lund, and Thomas~B Sch{\"o}n.
\newblock {Refusion: Enabling large-size realistic image restoration with latent-space diffusion models}.
\newblock In \emph{CVPR}, 2023{\natexlab{b}}.

\bibitem[Mallat(1989)]{192463}
S.G. Mallat.
\newblock {A theory for multiresolution signal decomposition: the wavelet representation}.
\newblock \emph{IEEE Transactions on Pattern Analysis and Machine Intelligence}, 11\penalty0 (7):\penalty0 674--693, 1989.

\bibitem[Pan et~al.(2017)Pan, Qiu, Yao, Li, and Mei]{pan2017create}
Yingwei Pan, Zhaofan Qiu, Ting Yao, Houqiang Li, and Tao Mei.
\newblock To create what you tell: Generating videos from captions.
\newblock In \emph{ACM Multimedia}, 2017.

\bibitem[Phung et~al.(2023)Phung, Dao, and Tran]{phung2023wavediff}
Hao Phung, Quan Dao, and Anh Tran.
\newblock {Wavelet Diffusion Models Are Fast and Scalable Image Generators}.
\newblock In \emph{CVPR}, 2023.

\bibitem[Rombach et~al.(2022{\natexlab{a}})Rombach, Blattmann, Lorenz, Esser, and Ommer]{ldm}
Robin Rombach, Andreas Blattmann, Dominik Lorenz, Patrick Esser, and Bj{\"o}rn Ommer.
\newblock {High-resolution image synthesis with latent diffusion models}.
\newblock In \emph{CVPR}, 2022{\natexlab{a}}.

\bibitem[Rombach et~al.(2022{\natexlab{b}})Rombach, Blattmann, Lorenz, Esser, and Ommer]{rombach2022high}
Robin Rombach, Andreas Blattmann, Dominik Lorenz, Patrick Esser, and Bj{\"o}rn Ommer.
\newblock {High-Resolution Image Synthesis With Latent Diffusion Models}.
\newblock In \emph{CVPR}, 2022{\natexlab{b}}.

\bibitem[Ruiz et~al.(2023)Ruiz, Li, Jampani, Pritch, Rubinstein, and Aberman]{ruiz2023dreambooth}
Nataniel Ruiz, Yuanzhen Li, Varun Jampani, Yael Pritch, Michael Rubinstein, and Kfir Aberman.
\newblock {DreamBooth: Fine tuning text-to-image diffusion models for subject-driven generation}.
\newblock In \emph{CVPR}, 2023.

\bibitem[Saharia et~al.(2022)Saharia, Chan, Saxena, Li, Whang, Denton, Ghasemipour, Gontijo~Lopes, Karagol~Ayan, Salimans, Ho, Fleet, and Norouzi]{NEURIPS2022_ec795aea}
Chitwan Saharia, William Chan, Saurabh Saxena, Lala Li, Jay Whang, Emily~L Denton, Kamyar Ghasemipour, Raphael Gontijo~Lopes, Burcu Karagol~Ayan, Tim Salimans, Jonathan Ho, David~J Fleet, and Mohammad Norouzi.
\newblock {Photorealistic Text-to-Image Diffusion Models with Deep Language Understanding}.
\newblock In \emph{NeurIPS}, 2022.

\bibitem[Schneider(2023)]{schneider2023archisound}
Flavio Schneider.
\newblock {Archisound: Audio Generation with Diffusion}.
\newblock \emph{arXiv preprint arXiv:2301.13267}, 2023.

\bibitem[Song et~al.(2021{\natexlab{a}})Song, Meng, and Ermon]{ddim}
Jiaming Song, Chenlin Meng, and Stefano Ermon.
\newblock {Denoising Diffusion Implicit Models}.
\newblock In \emph{ICLR}, 2021{\natexlab{a}}.

\bibitem[Song et~al.(2021{\natexlab{b}})Song, Sohl-Dickstein, Kingma, Kumar, Ermon, and Poole]{sde}
Yang Song, Jascha Sohl-Dickstein, Diederik~P Kingma, Abhishek Kumar, Stefano Ermon, and Ben Poole.
\newblock {Score-Based Generative Modeling through Stochastic Differential Equations}.
\newblock In \emph{ICLR}, 2021{\natexlab{b}}.

\bibitem[Stankovic and Falkowski(2003)]{stankovic2003haar}
Radomir~S Stankovic and Bogdan~J Falkowski.
\newblock {The Haar wavelet transform: its status and achievements}.
\newblock \emph{Computers \& Electrical Engineering}, 29\penalty0 (1):\penalty0 25--44, 2003.

\bibitem[Wang et~al.(2023)Wang, Dinh, Liu, and Xu]{wang2023boosting}
Xiyu Wang, Anh-Dung Dinh, Daochang Liu, and Chang Xu.
\newblock {Boosting diffusion models with an adaptive momentum sampler}.
\newblock \emph{arXiv preprint arXiv:2308.11941}, 2023.

\bibitem[Wells(1982)]{wells1982multirate}
Daniel~Raymond Wells.
\newblock \emph{{Multirate linear multistep methods for the solution of systems of ordinary differential equations}}.
\newblock University of Illinois at Urbana-Champaign, 1982.

\bibitem[Wizadwongsa et~al.(2023)Wizadwongsa, Chinchuthakun, Khungurn, Raj, and Suwajanakorn]{artifact}
Suttisak Wizadwongsa, Worameth Chinchuthakun, Pramook Khungurn, Amit Raj, and Supasorn Suwajanakorn.
\newblock {Diffusion Sampling with Momentum for Mitigating Divergence Artifacts}.
\newblock In \emph{ICLR}, 2023.

\bibitem[Wu et~al.(2023)Wu, Zhou, Kawaguchi, and Zhang]{wu2023fast}
Zike Wu, Pan Zhou, Kenji Kawaguchi, and Hanwang Zhang.
\newblock {Fast Diffusion Model}.
\newblock \emph{arXiv preprint arXiv:2306.06991}, 2023.

\bibitem[Xia et~al.(2023)Xia, Zhang, Wang, Wang, Wu, Tian, Yang, and Van~Gool]{xia2023diffir}
Bin Xia, Yulun Zhang, Shiyin Wang, Yitong Wang, Xinglong Wu, Yapeng Tian, Wenming Yang, and Luc Van~Gool.
\newblock {Diffir: Efficient diffusion model for image restoration}.
\newblock In \emph{ICLR}, 2023.

\bibitem[Yang et~al.(2023{\natexlab{a}})Yang, Chen, Pan, Yao, Chen, and Mei]{yang20233dstyle}
Haibo Yang, Yang Chen, Yingwei Pan, Ting Yao, Zhineng Chen, and Tao Mei.
\newblock 3dstyle-diffusion: Pursuing fine-grained text-driven 3d stylization with 2d diffusion models.
\newblock In \emph{ACM Multimedia}, 2023{\natexlab{a}}.

\bibitem[Yang et~al.(2022{\natexlab{a}})Yang, Wang, Chi, and Feng]{Yang2022WaveGAN}
Mengping Yang, Zhe Wang, Ziqiu Chi, and Wenyi Feng.
\newblock {WaveGAN: An Frequency-aware GAN for High-Fidelity Few-shot Image Generation}.
\newblock In \emph{ECCV}, 2022{\natexlab{a}}.

\bibitem[Yang et~al.(2022{\natexlab{b}})Yang, Wang, Chi, and Zhang]{yang2022fregan}
Mengping Yang, Zhe Wang, Ziqiu Chi, and Yanbing Zhang.
\newblock {FreGAN: Exploiting Frequency Components for Training GANs under Limited Data}.
\newblock In \emph{NeurIPS}, 2022{\natexlab{b}}.

\bibitem[Yang et~al.(2023{\natexlab{b}})Yang, Srivastava, and Mandt]{yang2023diffusion}
Ruihan Yang, Prakhar Srivastava, and Stephan Mandt.
\newblock {Diffusion Probabilistic Modeling for Video Generation}.
\newblock \emph{Entropy}, 25\penalty0 (10):\penalty0 1469, 2023{\natexlab{b}}.

\bibitem[Yang et~al.(2023{\natexlab{c}})Yang, Zhou, Feng, and Wang]{yang2023diffusionslim}
Xingyi Yang, Daquan Zhou, Jiashi Feng, and Xinchao Wang.
\newblock {Diffusion Probabilistic Model Made Slim}.
\newblock In \emph{CVPR}, 2023{\natexlab{c}}.

\bibitem[Yu et~al.(2015)Yu, Zhang, Song, Seff, and Xiao]{lsun}
Fisher Yu, Yinda Zhang, Shuran Song, Ari Seff, and Jianxiong Xiao.
\newblock {LSUN: Construction of a Large-scale Image Dataset using Deep Learning with Humans in the Loop}.
\newblock \emph{arXiv preprint arXiv:1506.03365}, 2015.

\bibitem[Yuan et~al.(2023)Yuan, Li, Wang, Yang, Lin, Liu, and Wang]{yuan2023spatial}
Xin Yuan, Linjie Li, Jianfeng Wang, Zhengyuan Yang, Kevin Lin, Zicheng Liu, and Lijuan Wang.
\newblock {Spatial-Frequency U-Net for Denoising Diffusion Probabilistic Models}.
\newblock \emph{arXiv preprint arXiv:2307.14648}, 2023.

\bibitem[Zhang et~al.(2022)Zhang, Gu, Zhang, Bao, Chen, Wen, Wang, and Guo]{zhang2022styleswin}
Bowen Zhang, Shuyang Gu, Bo Zhang, Jianmin Bao, Dong Chen, Fang Wen, Yong Wang, and Baining Guo.
\newblock {StyleSwin: Transformer-Based GAN for High-Resolution Image Generation}.
\newblock In \emph{CVPR}, 2022.

\bibitem[Zhang et~al.(2023{\natexlab{a}})Zhang, Rao, and Agrawala]{zhang2023adding}
Lvmin Zhang, Anyi Rao, and Maneesh Agrawala.
\newblock {Adding conditional control to text-to-image diffusion models}.
\newblock In \emph{ICCV}, 2023{\natexlab{a}}.

\bibitem[Zhang and Chen(2022)]{zhang2022fast}
Qinsheng Zhang and Yongxin Chen.
\newblock {Fast Sampling of Diffusion Models with Exponential Integrator}.
\newblock \emph{arXiv preprint arXiv:2204.13902}, 2022.

\bibitem[Zhang et~al.(2023{\natexlab{b}})Zhang, Tao, and Chen]{zhang2022gddim}
Qinsheng Zhang, Molei Tao, and Yongxin Chen.
\newblock {gDDIM: Generalized denoising diffusion implicit models}.
\newblock In \emph{ICLR}, 2023{\natexlab{b}}.

\bibitem[Zhao et~al.(2023)Zhao, Bai, Rao, Zhou, and Lu]{zhao2023unipc}
Wenliang Zhao, Lujia Bai, Yongming Rao, Jie Zhou, and Jiwen Lu.
\newblock {Uni{PC}: A Unified Predictor-Corrector Framework for Fast Sampling of Diffusion Models}.
\newblock In \emph{NeurIPS}, 2023.

\end{thebibliography}
	}
	
\end{document}